\documentclass[11pt]{article}

\usepackage[preprint]{acl}

\usepackage{times}
\usepackage{latexsym}

\usepackage[T1]{fontenc}

\usepackage[utf8]{inputenc}

\usepackage{microtype}

\usepackage{inconsolata}

\usepackage{tabularx}
\usepackage{graphicx}
\usepackage{booktabs}
\usepackage{multirow}
\usepackage{multicol}
\usepackage{amsmath}
\usepackage{mathtools}
\usepackage{tcolorbox}
\usepackage{listings}
\lstdefinestyle{plain}{
    basicstyle=\fontsize{7}{9.5}\ttfamily,
    keywordstyle=\color{blue},
    commentstyle=\color{gray},
    stringstyle=\color{green},
    showstringspaces=false,
    breaklines=true,
    breakatwhitespace=false,
    breakindent=0pt,
    escapeinside={(*@}{@*)}
}

\title{\textit{You Never Know a Person, You Only Know Their Defenses}: Detecting Levels of Psychological Defense Mechanisms in Supportive Conversations}

\author{
  Hongbin Na$^{1,}$\thanks{Equal contribution.}, Zimu Wang$^{2,}$\footnotemark[1], Zhaoming Chen$^{3,}$\footnotemark[1], \\ \textbf{Peilin Zhou$^{4}$, Yining Hua$^{5}$ Grace Ziqi Zhou$^{6}$, Haiyang Zhang$^{2}$, Tao Shen$^{1}$}, \\ \textbf{Wei Wang$^{2}$, John Torous$^{5}$, Shaoxiong Ji$^{7,8}$, Ling Chen$^{1}$} \\
  $^{1}$University of Technology Sydney \ \
  $^{2}$Xi'an Jiaotong-Liverpool University \ \
  $^{3}$University of Utah \\
  $^{4}$The Hong Kong University of Science and Technology (Guangzhou) \ \
  $^{5}$Harvard University \\
  $^{6}$The University of Sydney \ \
  $^{7}$ELLIS Institute Finland \ \
  $^{8}$University of Turku\\
}

\begin{document}
\maketitle
\begin{abstract}
Psychological defenses are strategies, often automatic, that people use to manage distress. Rigid or overuse of defenses is negatively linked to mental health and shapes what speakers disclose and how they accept or resist help. However, defenses are complex and difficult to reliably measure, particularly in clinical dialogues.  We introduce \textsc{PsyDefConv}, a dialogue corpus with help seeker utterances labeled for defense level, and \textsc{DMRS Co-Pilot}, a four-stage pipeline that provides evidence-based pre-annotations. The corpus contains 200 dialogues and 4{,}709 utterances, including 2{,}336 help seeker turns, with labeling and Cohen’s kappa 0.639. In a counterbalanced study, the co-pilot reduced average annotation time by 22.4\%. In expert review, it averaged 4.62 for evidence, 4.44 for clinical plausibility, and 4.40 for insight on a seven-point scale. Benchmarks with strong language models in zero-shot and fine-tuning settings demonstrate clear headroom, with the best macro F1-score around 30\% and a tendency to overpredict mature defenses. Corpus analyses confirm that mature defenses are most common and reveal emotion-specific deviations. We will release the corpus, annotations, code, and prompts to support research on defensive functioning in language.
\end{abstract}

\section{Introduction}
\begin{quote}
\textit{“The False Self, if successful in its function, hides the True Self.”}\\
\hfill---\citeauthor{winnicott2018ego}, \citeyear{winnicott2018ego}
\end{quote}
Emotional support conversation (ESC) builds dialogue agents that alleviate users' stress through multi-turn, strategy-grounded interaction, which is shaped not only by feelings but also by psychological defenses. Psychological defenses are among the oldest concepts in psychological science, and are defined as strategies people use to protect themselves from psychic pain, which in turn guides what they disclose, how they reframe difficulties, and how they accept or resist help~\cite{freud1936inhibitions}. Recent work has advanced affect modeling~\cite{wang-etal-2024-knowledge,wang-etal-2025-posts,ma-etal-2025-detecting,zhao2025hears}, empathy~\cite{hua2025charting, cai-etal-2024-empcrl,info:doi/10.2196/52597}, and strategy selection~\cite{kang-etal-2024-large,hua2025scoping,na-etal-2025-survey} for emotional support dialogue. While psychological defenses are important components of emotionally supportive conversations~\cite{di2024therapists}, the defensive function of utterances remains largely unmodeled in current ESC systems.

The Defense Mechanism Rating Scale (DMRS) is recognized as a gold standard taxonomy of defensive functioning, comprising three categories, seven levels, approximately thirty mechanisms, and 150 descriptive items validated in clinical research \cite{perry2004studying,HierarchyDMRSQ,vaillant1992ego,vaillant2012adaptation}. Its mechanism-level judgments are built for clinical case formulation that integrates patterns across time and situations. Conversational corpora show local linguistic behavior. Rather than treat this as a limitation, we align the unit of analysis with the unit of evidence. We operationalize DMRS levels, a sanctioned aggregation that captures defensive functioning from local cues. This choice enhances identifiability, inter-annotator agreement, and reproducibility while remaining faithful to the theory.


Despite the central role of defenses in clinical theory and practice, to our knowledge, there is no publicly available conversational dataset that annotates DMRS-based defensive functioning at any granularity. The absence of such a resource prevents a reproducible study of how defenses appear in language and blocks systematic evaluation of models on this construct.
To address this gap, we introduce \textsc{PsyDefConv}, a dialogue resource that labels defense levels for help seeker utterances. Building on ESConv \cite{liu-etal-2021-towards}, we draw a representative subset of 200 dialogues through stratified sampling over the joint distribution of problem types and emotions. The corpus contains 4{,}709 utterances in total with 2{,}336 seeker turns. To cover conversational phenomena, we augment the seven DMRS levels with two labels: \textit{No Defense} to mark phatic or functional turns, and \textit{Needs More Information} to mark cases where context is insufficient. Two trained annotators perform independent double blind labeling and reach substantial agreement with Cohen’s $\kappa=0.639$. Disagreements are adjudicated to form a gold standard.

To support efficiency and consistency, we introduce \textsc{DMRS Co-Pilot}, a four-stage pipeline that delivers level recommendations as pre-annotations. As shown in Figure \ref{fig:copilot_pipeline}, the system contextualizes the target text, screens candidate items, validates evidence, and synthesizes ranked conclusions. It is refined with annotator feedback and engineered for robust outputs under a strict schema. In a counterbalanced study, it reduces average time per task by 22.4\%. In expert review, it averages 4.62 for evidence, 4.44 for plausibility, and 4.40 for insight.

We conduct diverse experiments on a range of language models in both zero-shot and fine-tuning settings. Analyses show that mature defenses dominate overall. Shame displays a higher share of low-level defenses. Anger shows more utterances without defenses. Models tend to over-predict the high adaptive level, and the best macro F1-score is around 30\%, which indicates substantial headroom and the need for theory-aware supervision.

Our resource and method align clinical theory with what is observable in dialogue and address two practical challenges: the lack of a conversational dataset for defensive functioning and the under-specification of mechanism labels in text-only episodes. By focusing on DMRS levels and tooling the workflow, we provide a reproducible basis for study and evaluation. Overall, our main contributions are as follows:
(1) \emph{Resource}: to the best of our knowledge, we release \textsc{PsyDefConv}, the first conversational dataset annotated with DMRS defense levels, covering 2{,}336 help seeker utterances with double-blind labeling and substantial agreement ($\kappa=0.639$);
(2) \emph{Tooling}: we present \textsc{DMRS Co-Pilot}, a four-stage pipeline that contextualizes the text, screens and validates candidate items, and synthesizes ranked level recommendations as pre-annotations to support consistency and efficiency;
(3) \emph{Benchmark and analysis}: we evaluate strong language models in zero-shot and fine-tuning settings, and report error structures and distributional findings, with best macro F1-score around 30\%, establishing headroom and guiding future work.

\section{\textsc{PsyDefConv}}

\subsection{Data Source}
We build upon ESConv~\cite{liu-etal-2021-towards}, a large-scale, publicly English corpus of emotional support dialogues, and follow its role definitions of seeker and supporter. To obtain a representative 200-dialogue subset, we perform stratified sampling on the joint distribution of problem and emotion types observed in ESConv. We allocate quotas to each combination of the five major problem types and six core emotions in proportion to their ESConv frequencies, thereby preserving the source joint distribution and supporting a broad coverage of dialogue types.

\subsection{Annotation Scheme}
Each instance corresponds to a single seeker utterance at the utterance level. We focus solely on annotating seeker turns to capture the defensive responses of the distressed individual, rather than the strategies employed by the supporter. Annotators have access only to the dialogue context preceding and including the current seeker utterance. They do not see subsequent turns and annotate only the current utterance. The annotators are two fluent English speakers with expertise in both psychology and natural language processing.

Our scheme adapts the Defense Mechanism Rating Scales (DMRS)~\cite{perry2004studying,HierarchyDMRSQ}, an empirically validated rating scale that operationalizes defenses into a hierarchy comprising three categories, seven defense levels, and approximately thirty mechanisms, defined via 150 descriptive items. Although DMRS targets mechanism-level assessment in clinical settings, applying mechanisms to dialogue transcripts is difficult due to the lack of longitudinal, cross-situational evidence. We therefore label \emph{defense levels}, which provide functionally coherent, semantically stable groupings that are more reliably identifiable from dialogue text, supporting higher inter-annotator agreement and reproducibility.

To accommodate conversational phenomena, we add two labels: utterances without defensive function (e.g., greetings, phatic expressions) are labeled \textit{No Defense}, and utterances that are unclassifiable due to insufficient context are labeled \textit{Needs More Information}. These extensions increase coverage of non-defensive and ambiguous cases commonly observed in natural dialogues and reduce annotation noise. For an overview of the different defense levels and their corresponding mechanisms, refer to Table~\ref{tab:defense_overview_compact}.

\begin{table*}[t]
\centering
\scriptsize
\renewcommand{\arraystretch}{0.95}

\begin{tabularx}{\textwidth}{@{} l >{\raggedright\arraybackslash}X @{}}
\toprule
\multicolumn{2}{@{}l}{\textbf{Level 0: No Defenses}} \\
\midrule
\textit{Mechanisms} & N/A \\
\textit{Definition} & Purely functional utterances that serve to maintain conversational flow, express social niceties, or exchange non-emotional information. They do not engage with psychological conflict. \\
\textit{Example} & Hello, thank you for your time, or a simple Okay. \\
\midrule

\multicolumn{2}{@{}l}{\textbf{Level 1: Action Defenses}} \\
\midrule
\textit{Mechanisms} & Passive Aggression, Help-Rejecting Complaining, Acting Out \\
\textit{Definition} & Dealing with internal conflicts or external stressors by acting on the environment. The individual's distress is channeled into behavior, often impulsively and without reflection, as a way to release tension, gratify wishes, or avoid fears and painful feelings. \\
\textit{Example} & After being criticized at work, instead of discussing the issue, a person goes home and starts a fight with their partner over something minor. \\
\midrule

\multicolumn{2}{@{}l}{\textbf{Level 2: Major Image-Distorting Defenses}} \\
\midrule
\textit{Mechanisms} & Splitting (of self-image and others' image), Projective Identification \\
\textit{Definition} & Coping with intolerable anxiety by grossly distorting the image of oneself or others. This is achieved by splitting representations into polar opposites (all-good or all-bad), which simplifies reality and protects the individual from the anxiety of dealing with ambivalence. \\
\textit{Example} & “My new boss is a genius and will solve everything,” or "My boss is completely incompetent and is ruining the company." \\
\midrule

\multicolumn{2}{@{}l}{\textbf{Level 3: Disavowal Defenses}} \\
\midrule
\textit{Mechanisms} & Denial, Rationalization, Projection, Autistic Fantasy \\
\textit{Definition} & Dealing with stressors by refusing to acknowledge unacceptable aspects of reality or one's own experience. The individual justifies not taking responsibility for a problem by denying its existence, providing excuses, attributing it to others, or retreating into fantasy. \\
\textit{Example} & "I didn't get the promotion because my boss is biased against me, not because my performance was lacking." \\
\midrule

\multicolumn{2}{@{}l}{\textbf{Level 4: Minor Image-Distorting Defenses}} \\
\midrule
\textit{Mechanisms} & Devaluation (of self-image and others' image), Idealization (of self-image and others' image), Omnipotence \\
\textit{Definition} & Protecting self-esteem from threats like failure or criticism by distorting one's image in a less severe manner than Level 2. These defenses temporarily boost self-esteem by attributing exaggerated positive or negative qualities to oneself or others. \\
\textit{Example} & "Sure, they succeeded, but it was just luck. Anyone could have done it." \\
\midrule

\multicolumn{2}{@{}l}{\textbf{Level 5: Neurotic Defenses}} \\
\midrule
\textit{Mechanisms} & Repression, Dissociation, Reaction Formation, Displacement \\
\textit{Definition} & Managing emotional conflict by keeping unacceptable wishes, thoughts, or motives out of conscious awareness. The individual may experience the feelings associated with a conflict while the idea is blocked (or vice versa), leading to indirect or displaced expressions. \\
\textit{Example} & Feeling unexplainably irritable and anxious all day, only to later realize it's the anniversary of a forgotten painful event. \\
\midrule

\multicolumn{2}{@{}l}{\textbf{Level 6: Obsessional Defenses}} \\
\midrule
\textit{Mechanisms} & Isolation of Affect, Intellectualization, Undoing \\
\textit{Definition} & Managing threatening feelings by separating them from the thoughts or events that caused them. The individual remains aware of the cognitive details but avoids the emotional impact by using excessive logic, abstract thinking, or symbolic acts to maintain control. \\
\textit{Example} & Describing a traumatic car accident with precise, technical detail but showing no emotion, as if reading a report. \\
\midrule

\multicolumn{2}{@{}l}{\textbf{Level 7: High-Adaptive Defenses}} \\
\midrule
\textit{Mechanisms} & Affiliation, Altruism, Anticipation, Humor, Self-Assertion, Self-Observation, Sublimation, Suppression \\
\textit{Definition} & Representing the most adaptive and constructive ways of handling stressors. The individual faces conflict by consciously integrating feelings with thoughts, anticipating challenges, seeking support, and channeling emotions into productive outcomes. \\
\textit{Example} & "I'm feeling overwhelmed by this project. I'm going to call a colleague to talk through some strategies and get a fresh perspective." \\
\midrule

\multicolumn{2}{@{}l}{\textbf{Level 8: Needs More Information}} \\
\midrule
\textit{Mechanisms} & N/A \\
\textit{Definition} & Used when an utterance is too ambiguous or the context is insufficient. \\
\textit{Example} & When a user's reply is a vague "maybe" in response to a deep emotional question, the intent cannot be determined. \\
\bottomrule
\end{tabularx}
\caption{Overview of defense mechanisms and their corresponding levels, including examples and descriptions.}
\label{tab:defense_overview_compact}
\vspace{-2mm}
\end{table*}

\subsection{Annotator Training} 
Before full-scale annotation, we ran a structured annotator training program to validate the labeling scheme and iteratively refine operational guidelines. The program comprised four iterations, each covering 10 dialogues, for a total of 461 seeker utterances. Two annotators independently labeled every utterance using our nine-category scheme; after each iteration, we held calibration meetings to review disagreements and update the codebook.

The agreement between annotators progressed over time. Iteration~1 yielded moderate reliability ($\kappa=0.496$). After we introduced a simplified evaluation protocol, reliability dipped in Iteration~2 ($\kappa=0.426$) but then rose sharply: Iteration~3 reached $\kappa=0.642$, and Iteration~4 $\kappa=0.667$, indicating a shift from moderate to substantial agreement. Overall, reliability improved by 34.5\% from the first to the final iteration.



\subsection{Formal Annotation}
The formal annotation of all 200 dialogues, consisting of 2,336 Seeker utterances, was conducted by two trained annotators using the Label Studio \cite{LabelStudio} platform developed by HumanSignal\footnote{\url{https://humansignal.com/}}. Analysis results from the \emph{DMRS Co-Pilot} system were imported as pre-annotation suggestions to support efficiency. Independent double-blind annotation yielded a Cohen’s $\kappa$ of 0.639, indicating substantial agreement. To construct the final gold-standard corpus, disagreements were reviewed jointly and resolved by consensus with reference to the annotation handbook. 

\subsection{Dataset Statistics}

As shown in Table~\ref{tab:basic_stats}, 
\textsc{PsyDefConv} consists of 200 dialogues and 4,709 utterances evenly split between seekers and supporters. Each dialogue contains an average of 23.5 turns, with a mean utterance length of 19.8 tokens.

\paragraph{Distribution of Psychological Attributes.}
Table~\ref{tab:psydef_distributions} presents the distribution of psychological factors annotated at the seeker level, including presenting problems, expressed emotions, and defense mechanisms. The most common problems include \textit{Ongoing Depression} and \textit{Job Crisis}, while \textit{Anxiety}, \textit{Depression}, and \textit{Sadness} emerge as the dominant emotional expressions. In terms of defense mechanisms, mature defense strategies are the most frequently observed (over 50\%), while neurotic and immature defenses are less common. Notably, 17.4\% of utterances are either non-defensive or contextually ambiguous, reflecting emotionally neutral or under-specified dialogue turns.

\begin{table}[t]
\centering
\resizebox{\columnwidth}{!}{%
\begin{tabular}{lrrr}
\toprule
\textbf{Category} & \textbf{Total} & \textbf{Supporter} & \textbf{Seeker} \\
\midrule
\# Dialogues & 200 & -- & -- \\
\# Utterances & 4,709 & 2,373 & 2,336 \\
Avg. Turns per Dialogue & 23.5 $\pm$ 6.6 & 11.9 $\pm$ 3.4 & 11.7 $\pm$ 3.3 \\
Avg. Length of Utterances & 19.8 $\pm$ 16.5 & 20.9 $\pm$ 17.0 & 18.8 $\pm$ 15.8 \\
\bottomrule
\end{tabular}
}
\caption{Data Statistics of \textsc{PsyDefConv}.}
\label{tab:basic_stats}
\vspace{-4mm}
\end{table}

\begin{figure}[t]
    \centering
    \includegraphics[width=1\linewidth]{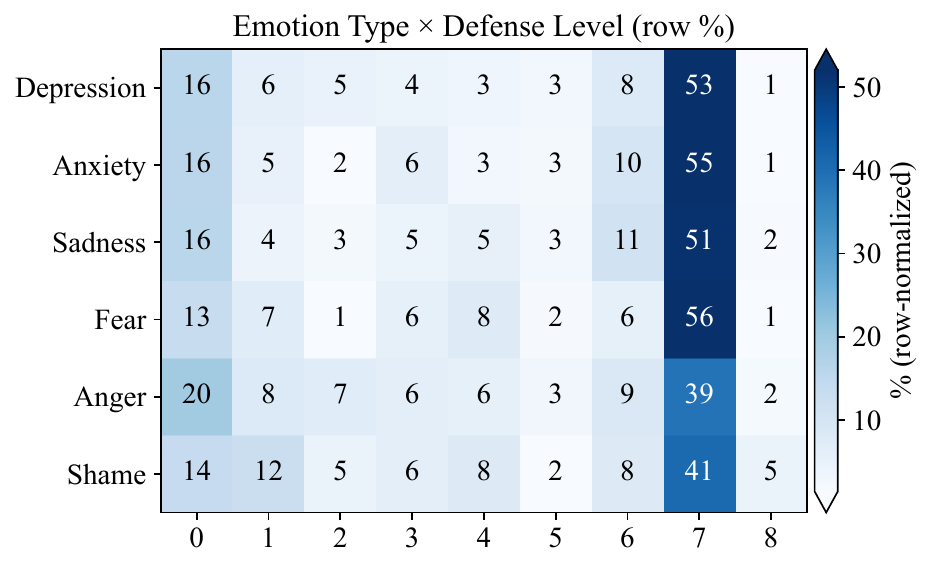}
    \caption{Distribution of defense levels (x-axis) across different emotions (y-axis). 
Values indicate the proportion of each defense level within a given emotion.}
    \label{fig:dis_defense_levels}
\end{figure}

\begin{figure}[t]
    \centering
    \includegraphics[width=0.95\linewidth]{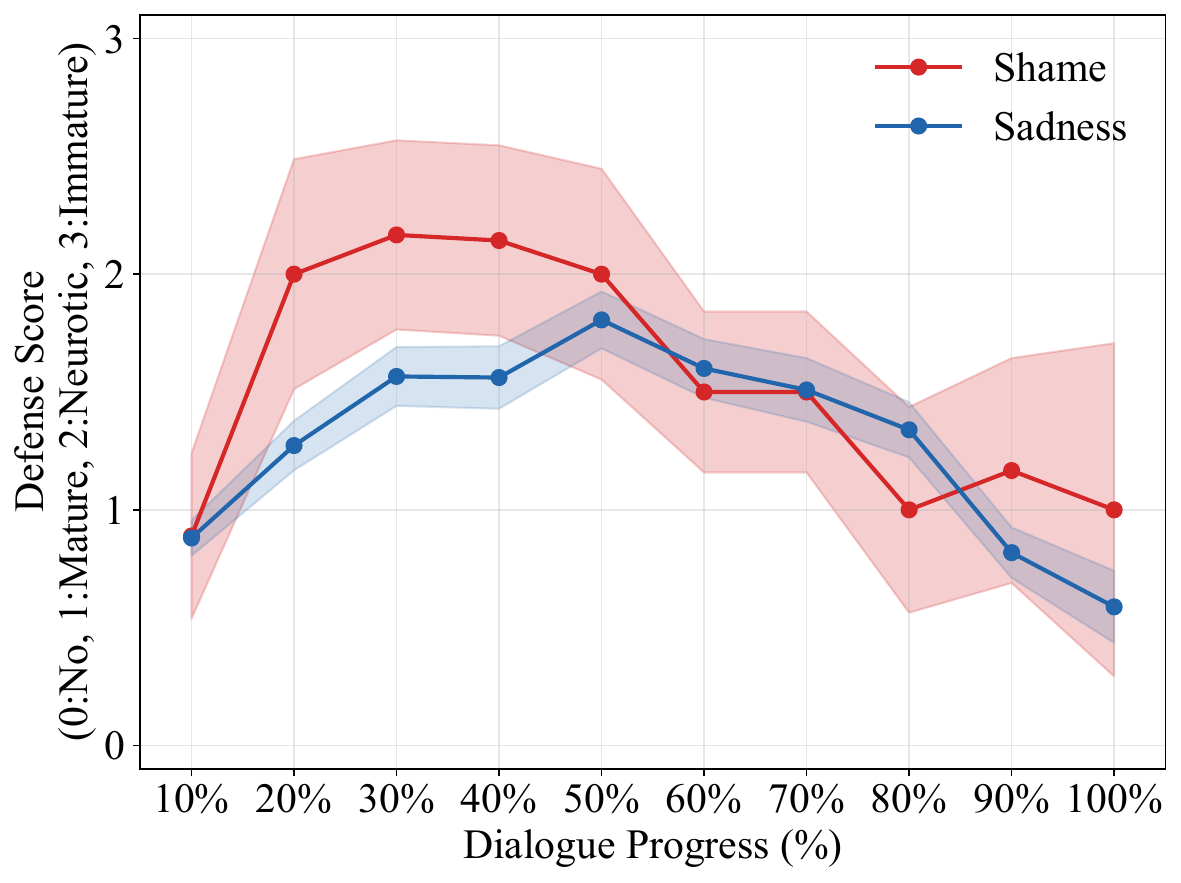}
    \caption{
Average defense scores over dialogue progress for seeker utterances expressing \textit{shame} and \textit{sadness}. 
Shame shows stronger early defensiveness and greater fluctuation than sadness.
}
    \vspace{-4mm}
    \label{fig:intro}
\end{figure}

\paragraph{Distribution of Defense Levels Across Emotions.}
Figure~\ref{fig:dis_defense_levels} shows that mature defenses (Level~7) dominate across all emotions, indicating a general preference for adaptive coping. However, variations emerge across emotion types. \textit{Shame} exhibits the highest proportion of low-level defenses (Levels~1–4), suggesting more fragmented and immature regulation. \textit{Anger} stands out for its elevated use of Level~0 (no defense), implying more unfiltered or confrontational expressions. In contrast, \textit{depression}, \textit{anxiety}, and \textit{sadness} exhibit highly similar distributions across all nine defense levels, reflecting a consistent and balanced defensive response.

\paragraph{Emotion-Specific Trajectories.}
Figure~\ref{fig:intro} illustrates how defense strategies evolve over the course of dialogues for seeker utterances expressing \textit{shame} and \textit{sadness}. While both emotions initially rise from non-defensive to more regulated responses, shame triggers a sharper early increase in neurotic defenses, followed by a steeper decline. In contrast, sadness exhibits a flatter trajectory centered around mature defenses. These patterns suggest that shame is associated with more reactive and volatile defense regulation compared to the steadier coping observed in sadness.


\begin{table}[t]
  \centering
  \footnotesize
  \resizebox{\columnwidth}{!}{%
    \begin{tabular}{lrr}
      \toprule
      \textbf{Categories} & \textbf{Num} & \textbf{Proportion} \\
      \midrule
      \multicolumn{3}{c}{\textbf{Seeker’s Problems}} \\
      \midrule
      Ongoing Depression      & 58    & 29.0\% \\
      Job Crisis              & 49    & 24.5\% \\
      Breakup with Partner    & 42    & 21.0\% \\
      Problems with Friends   & 26    & 13.0\% \\
      Academic Pressure       & 25    & 12.5\% \\
      \midrule
      Overall                 & 200   & 100.0\% \\
      \midrule
      \multicolumn{3}{c}{\textbf{Seeker’s Emotions}} \\
      \midrule
      Anxiety                 & 57    & 28.5\% \\
      Depression              & 55    & 27.5\% \\
      Sadness                 & 50    & 25.0\% \\
      Fear                    & 18    & 9.0\% \\
      Anger                   & 15    & 7.5\% \\
      Shame                   & 5     & 2.5\% \\
      \midrule
      Overall                 & 200   & 100.0\% \\
      \midrule
      \multicolumn{3}{c}{\textbf{Seeker’s Defense Levels}} \\
      \midrule
      0 No Defense                      & 371   & 15.9\% \\
      1 Action Level                 & 136   & 5.8\% \\
      2 Major Image-Distorting Level          & 77    & 3.3\% \\
      3 Disavowal Level              & 124   & 5.3\% \\
      4 Minor Image-Distorting Level          & 105   & 4.5\% \\
      5 Neurotic Level               & 61    & 2.6\% \\
      6 Obsessional Level            & 216   & 9.2\% \\
      7 High-Adaptive Level          & 1,211 & 51.8\% \\
      8 Needs More Information          & 35    & 1.5\% \\
      \midrule
      Overall                              & 2,336 & 100.0\% \\
      \midrule
      \multicolumn{3}{c}{\textbf{Seeker’s Defense Categories}} \\
      \midrule
      No Defense (0, 8)                   & 406   & 17.4\% \\
      Mature Defenses (7)          & 1,211 & 51.8\% \\
      Neurotic Defenses (5, 6)            & 277   & 11.9\% \\
      Immature Defenses (1, 2, 3, 4)      & 442   & 18.9\% \\
      \midrule
      Overall                              & 2,336 & 100.0\% \\
      \bottomrule
    \end{tabular}%
  }
  \caption{Distribution of the \textsc{PsyDefConv} dataset across seekers’ presenting problems, expressed emotions, annotated defense levels, and aggregated defense categories.
}
  \label{tab:psydef_distributions}
  \vspace{-2mm}
\end{table}

\section{\textsc{DMRS Co-Pilot}}
\begin{figure}[t!]
    \centering
    \includegraphics[width=\linewidth]{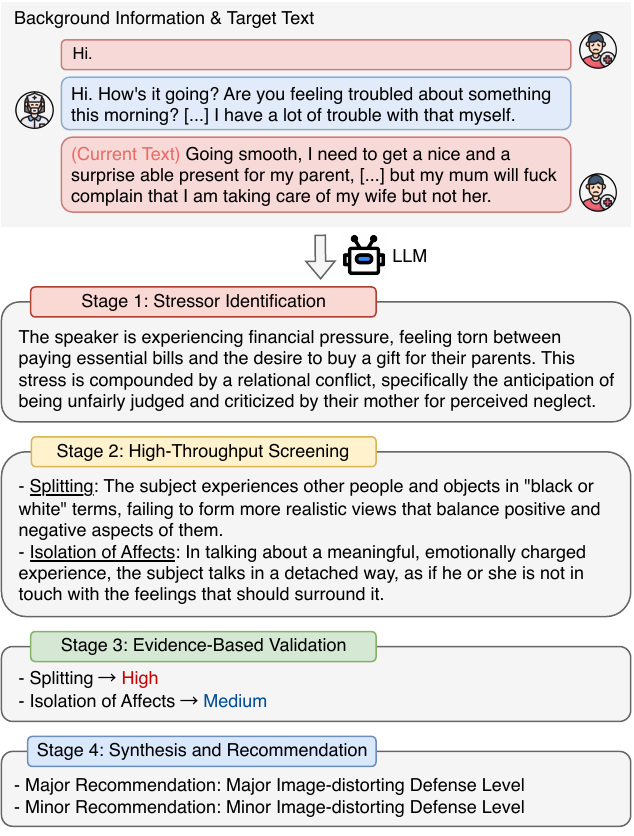}
    \caption{The four-stage analysis pipeline of \textsc{DMRS Co-Pilot}. The system follows a cascaded design, moving from broad contextualization and screening to deep validation and synthesis, allowing for the strategic use of different models for tasks with varying computational and reasoning requirements.}
    \label{fig:copilot_pipeline}
    \vspace{-2mm}
\end{figure}

To support annotation, we develop \textsc{DMRS Co-Pilot}, an automated system that operationalizes the DMRS framework to analyze target text. We iteratively refine the system across annotator training iterations, incorporating feedback from annotators and calibration outcomes. As shown in Figure~\ref{fig:copilot_pipeline}, the system implements a four-stage pipeline that contextualizes the target text using the background information, filters candidate defense mechanisms, validates the candidates, and synthesizes evidence into ranked recommendations. These recommendations are delivered as pre-annotation suggestions, improving efficiency and consistency.

\subsection{System Architecture}
\label{sec:copilot_pipeline}

We formalize the pipeline as a function $\Phi$ that maps the \emph{target text} and its \emph{background information} to a ranked pair of analytical conclusions.
Let the background information be $B$, and the target text be $x_t$.
The input to the pipeline is $(B, x_t)$; the output is a ranked pair $(\mathcal{C}_1, \mathcal{C}_2)$, where each conclusion specifies a concrete defense level from the predefined label set and provides a brief rationale that summarizes supporting evidence from the background information and the target text. The process comprises four sequential stages. We define
\begin{equation}
  \Phi \coloneqq f_4 \circ f_3 \circ f_2 \circ f_1 .
\end{equation}
Applying $\Phi$ to $(B, x_t)$ yields
\begin{equation}
  (\mathcal{C}_1, \mathcal{C}_2) = \Phi(B, x_t) .
\end{equation}

\paragraph{Stage 1: Stressor Identification ($f_1$).}
The first stage contextualizes the analysis by conditioning on the background information and the target text, and then identifies a candidate psychological stressor. 
The function $f_1$ takes $(B, x_t)$ as input and outputs a natural-language hypothesis $s$:
\begin{equation}
  s = f_1(B, x_t) .
\end{equation}

\paragraph{Stage 2: High-Throughput Screening ($f_2$).}
Let $\mathcal{I}=\{I_1,\dots,I_{150}\}$ denote the full set of DMRS descriptive items.
The screening stage reduces this set to a compact candidate subset by assessing each item’s relevance to $(B, x_t)$ conditioned on the hypothesized stressor $s$.
The function $f_2$ takes $(B, x_t, s)$ as input and returns, for each item, a binary relevance judgment and a brief rationale explaining the decision, from which the relevant subset $\mathcal{I}_{\mathrm{rel}}$ is derived:
\begin{equation}
  \mathcal{I}_{\mathrm{rel}} = f_2(B, x_t, s), \qquad \mathcal{I}_{\mathrm{rel}} \subseteq \mathcal{I}.
\end{equation}

\paragraph{Stage 3: Evidence-Based Validation ($f_3$).}
For each relevant item $I_j \in \mathcal{I}_{\mathrm{rel}}$, the function $f_3$ performs in-depth validation against $(B, x_t)$, conditioned on the hypothesized stressor $s$.
Each validated item yields a structured tuple $V_j = (c_j, e_j)$, where $c_j \in \{\text{High}, \text{Medium}, \text{Low}\}$ is a categorical confidence score and $e_j$ is textual evidence summarizing the support in $(B, x_t)$.
The overall output is the set of validated pairs:
\begin{equation}
  \mathcal{V} = f_3(\mathcal{I}_{\mathrm{rel}}, B, x_t, s).
\end{equation}

\paragraph{Stage 4: Synthesis and Recommendation ($f_4$).}
The final stage synthesizes $V$ together with the contextual information $(B, x_t)$ and the hypothesized stressor $s$ to produce two ranked analytical conclusions.
The function $f_4$ takes $(V, B, x_t, s)$ as input and outputs a primary and a secondary conclusion; each conclusion consists of a selected item, a defense-level label from the predefined label set, a concise rationale, and salient relational cues grounded in $(B, x_t)$:
\begin{equation}
  (\mathcal{C}_1, \mathcal{C}_2) = f_4(V, B, x_t, s).
\end{equation}

\subsection{Implementation Details}
The pipeline’s conceptual functions are implemented with large language models (LLMs), with model choice determined by each stage’s needs. Functions requiring deep contextual reasoning and synthesis (\(f_1, f_3, f_4\)) use Gemini 2.5 Pro, while the high-throughput screening function (\(f_2\)) uses the more efficient Gemini 2.5 Flash for scalability. To support reproducibility and reliability, all prompts run with a low temperature of 0.2, and the outputs of \(f_2\), \(f_3\), and \(f_4\) conform to a strict JSON schema for machine readability. The pipeline is engineered for robustness with parallel execution for \(f_2\) and \(f_3\), automated API call retries, and structured fallbacks for error handling.

\subsection{System Evaluation}
\paragraph{Annotator Evaluation of DMRS Co-Pilot.}
Table~\ref{tab:copilot_eval} summarizes two annotator-focused tests. 
Panel~A reports perceived usefulness as the share of utterances that both annotators marked helpful. 
Rates were 58.1\% in Iteration~1, 54.1\% in Iteration~2, 60.2\% in Iteration~3, and 67.3\% in Iteration~4, representing an overall increase of 9.2 percentage points. 
Panel~B measures efficiency in a within-subject counterbalanced crossover with a newly recruited annotator and a constant handbook. 
Group~1 moved from 33.10\,s without the tool to 19.66\,s with it, a 40.6\% speed up. 
Group~2 remained at 22.94\,s in both conditions. 
Across 126 tasks, the average time per task was 27.62\,s without and 21.43\,s with, a 22.4\% gain. 
Counterbalancing reduces practice and fatigue. 
Comparing the first blocks under each condition shows a 30.7\% advantage when the annotator starts with \textsc{DMRS Co-Pilot}.

\begin{table}[t]
\centering
\resizebox{\columnwidth}{!}{%
\begin{tabular}{lcccc}
\toprule
\multicolumn{5}{l}{\textbf{Panel A: Perceived usefulness (both annotators rated \emph{helpful})}}\\
\midrule
Metric & Iter.~1 & Iter.~2 & Iter.~3 & Iter.~4 \\
Helpful rate (\%) & 58.1 & 54.1 & 60.2 & 67.3 \\
\midrule[1.2pt] 
\multicolumn{5}{l}{\textbf{Panel B: Counterbalanced efficiency (within-subject crossover, same handbook)}}\\
\midrule
Group & Condition & Time/task (s) & \multicolumn{2}{c}{Speed-up vs.\ -} \\
\midrule
G1 & -             & 33.10 & \multicolumn{2}{c}{---} \\
G1 & DMRS Co-Pilot & 19.66 & \multicolumn{2}{c}{+40.6\%} \\
G2 & -             & 22.94 & \multicolumn{2}{c}{---} \\
G2 & DMRS Co-Pilot & 22.94 & \multicolumn{2}{c}{+0.0\%} \\
Pooled & -             & 27.62 & \multicolumn{2}{c}{---} \\
Pooled & DMRS Co-Pilot & 21.43 & \multicolumn{2}{c}{+22.4\%} \\
\bottomrule
\end{tabular}%
}
\vspace{4pt}
\caption{Perceived usefulness (Panel~A) and counterbalanced efficiency (Panel~B) of DMRS Co-Pilot. Notes: helpful rates for Iter.~2/3 are not reported here (replace “---” when available).}
\label{tab:copilot_eval}
\vspace{-4mm}
\end{table}

\paragraph{Expert Evaluation of DMRS Co-Pilot.}
Two expert reviewers, a professor of psychiatry and a doctoral student, independently assessed 45 system analyses produced by \textsc{DMRS Co-Pilot}. For each target utterance, they saw the dialogue context and the system’s hypothesized stressor, conclusions, and validated items, then rated \emph{Evidence Support}, \emph{Insightfulness}, and \emph{Clinical Plausibility} on a seven-point scale. Across all items and both reviewers, the system scored 4.62 for evidence, 4.44 for plausibility, and 4.40 for insight, with medians of 5 on all three dimensions, indicating generally adequate support and clinical fit. Agreement patterns were mixed. Insight showed the tightest alignment between reviewers, with a mean absolute difference of 1.87 and 51\% of cases within one point. Plausibility and evidence had mean absolute differences of 2.22 and 2.09, with 38\% and 42\% within one point. In 36\% of items, all three dimensions were within one point, while in 42\%, at least one dimension differed by three points or more, reflecting expected differences in expert judgement for brief conversational evidence. The system most often earned credit for citing sufficient textual evidence and was more variable in generating novel insights.

\section{Experiments}

\subsection{Experimental Setup}

We conduct experiments in both zero-shot and fine-tuning settings using state-of-the-art LLMs.
For zero-shot prompting, we experiment with GPT-5\footnote{\url{https://openai.com/index/gpt-5-system-card/}} (\texttt{2025-08-07}), GPT-5 mini (\texttt{2025-08-07}), Gemini 2.5 Pro \cite{comanici2025gemini25}, Kimi-K2 (\texttt{0905}, \citealp{kimiteam2025kimik2}), DeepSeek-V3.2 \cite{deepseekai2025deepseekv3}, and Qwen3-Next \cite{yang2025qwen3}, where the default/recommended hyperparameters are adopted. We also fine-tune models including Llama 3.1-8B \cite{grattafiori2024llama3}, Ministral-8B\footnote{\url{https://mistral.ai/news/ministraux}}, GLM-4-9B \cite{glm2024chatglm}, Qwen3-4B, Qwen3-8B, and InternLM3-8B \cite{cai2024internlm2}.
During the fine-tuning process, we set the number of epochs to 10, the batch size to 1, the gradient accumulation step to 8, and the learning rate to 1e-4.
Performance is assessed using accuracy, along with macro precision, recall, and F1-scores, evaluated on the positive classes (1-8).

\begin{table}[t!]
    \centering
    \small
    \begin{tabular}{l|cccc}
        \toprule
        \textbf{Model} & \textbf{ACC} & \textbf{P} & \textbf{R} & \textbf{F1} \\
        \midrule
        \multicolumn{5}{c}{\textbf{Zero-shot Prompted LLMs}} \\
        \midrule
        GPT-5 & 52.75 & \underline{27.59} & 16.56 & 19.53 \\ 
        GPT-5 mini & 54.03 & 26.30 & 16.99 & 18.41 \\
        Gemini 2.5 Pro & \textbf{56.36} & 27.49 & \underline{26.12} & \underline{25.99} \\
        Kimi-K2 & 41.10 & 21.28 & 21.38 & 17.58 \\
        DeepSeek-V3.2 (w/o) & 39.83 & 24.77 & 19.97 & 16.15 \\
        DeepSeek-V3.2 (w/) & \underline{55.72} & \textbf{29.66} & \textbf{27.53} & \textbf{26.17} \\
        Qwen3-Next (w/o) & 40.89 & 26.91 & 19.09 & 14.88 \\
        Qwen3-Next (w/) & 44.28 & 27.23 & 21.58 & 20.68 \\
        \midrule
        \multicolumn{5}{c}{\textbf{Fine-tuned LLMs}} \\
        \midrule
        Llama3.1-8B & 62.92 & 33.24 & \underline{30.08} & 30.51 \\
        Ministral-8B & \textbf{64.83} & \textbf{33.97} & \textbf{30.45} & \textbf{31.48} \\
        GLM-4-9B & 62.92 & 30.10 & 29.53 & 28.61 \\
        Qwen3-4B & 60.59 & 30.49 & 28.53 & 28.46 \\
        Qwen3-8B & 61.44 & 30.10 & 28.91 & 28.39 \\
        InternLM3-8B & \underline{63.98} & \underline{33.47} & 29.93 & \underline{30.53} \\
        \bottomrule
    \end{tabular}
    \caption{Overall performance of the evaluated models on the \textsc{PsyDefConv} dataset. ``w/o'' and ``w/'' denote the thinking process is disabled/enabled. The best and the second-best performance in each setup are highlighted in \textbf{bold} and \underline{underlined}, respectively.}
    \label{tab:overall}
    \vspace{-4mm}
\end{table}

\begin{figure}[t]
    \centering
    \includegraphics[width=\linewidth]{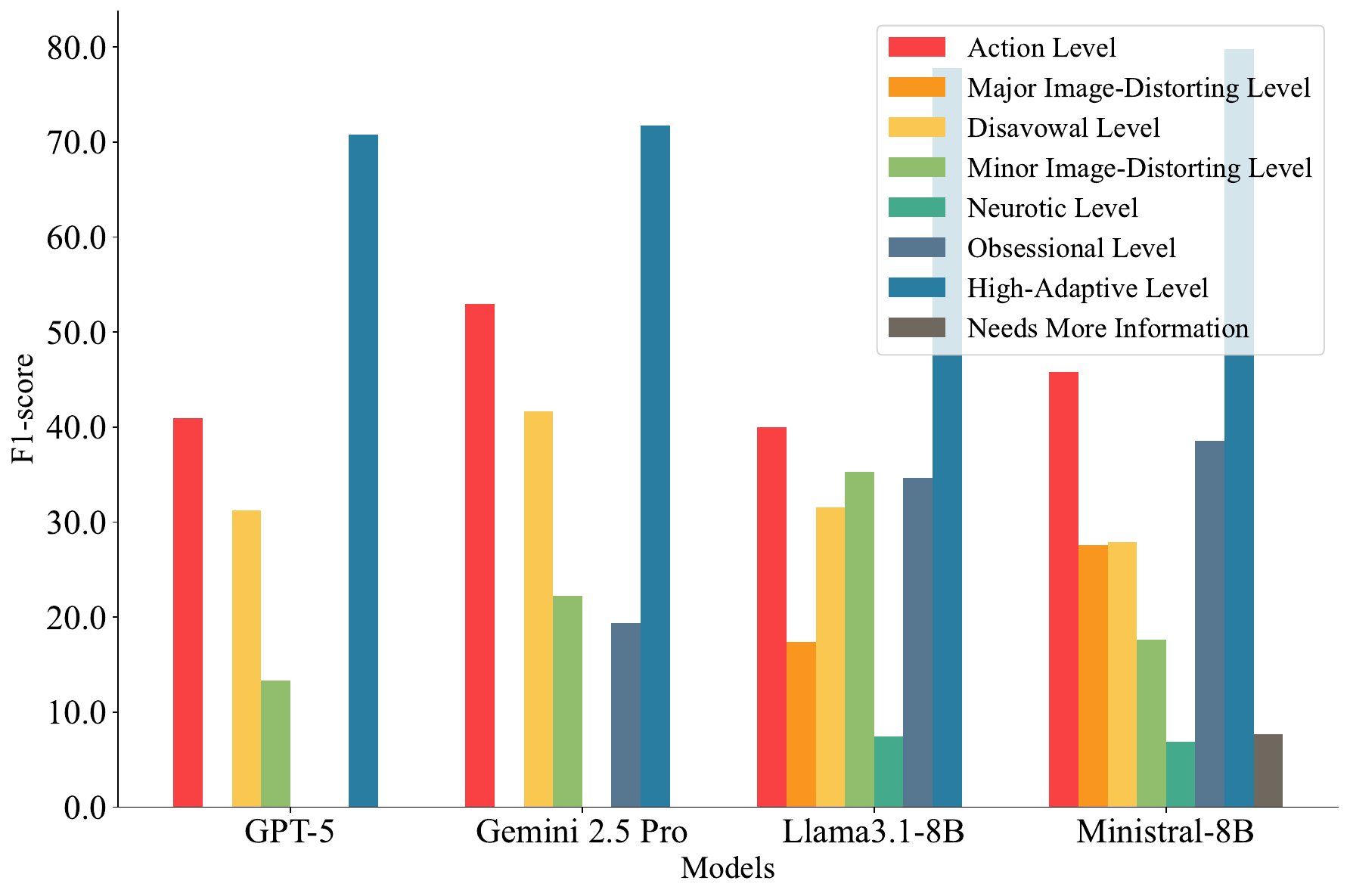}
    \caption{F1-score of the models with respect to different levels of psychological defense mechanisms.}
    \label{fig:per-class}
    \vspace{-4mm}
\end{figure}

\begin{figure}[t]
    \centering
    \includegraphics[width=\linewidth]{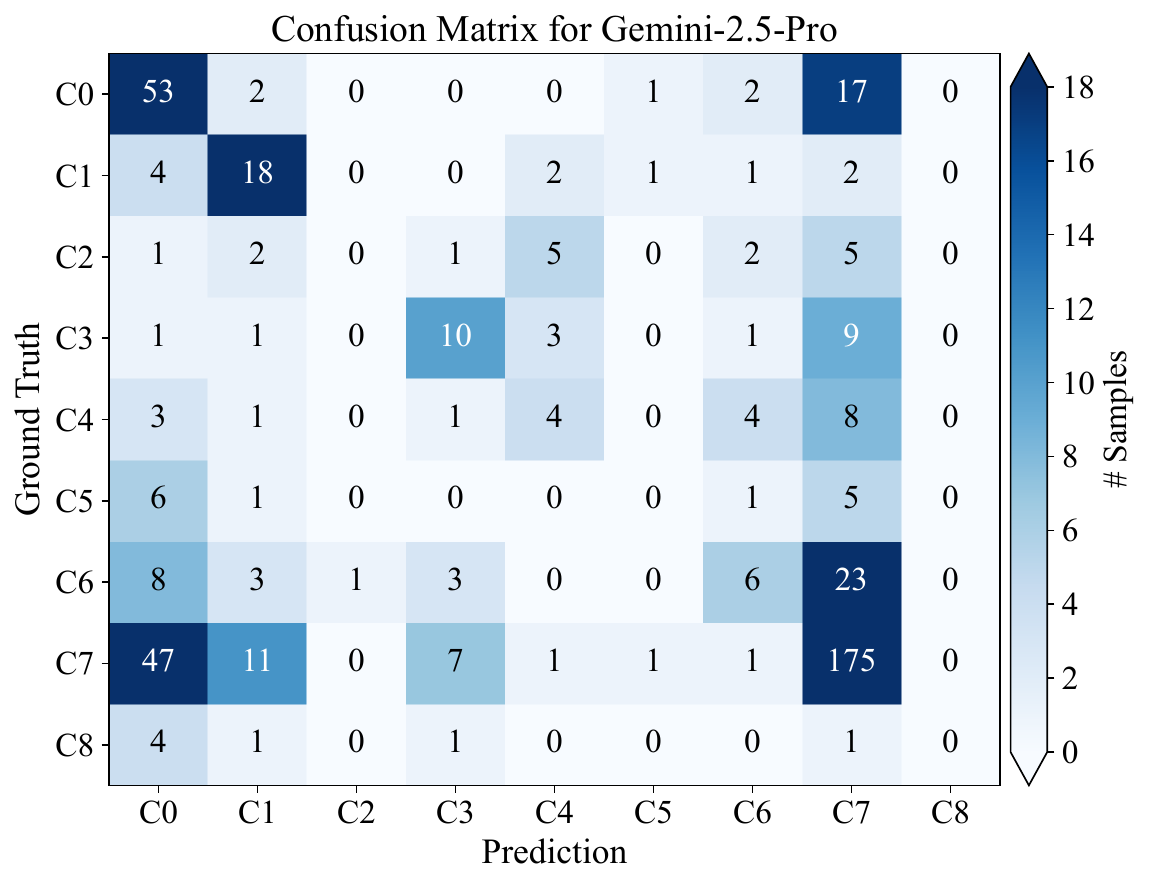}
    \includegraphics[width=\linewidth]{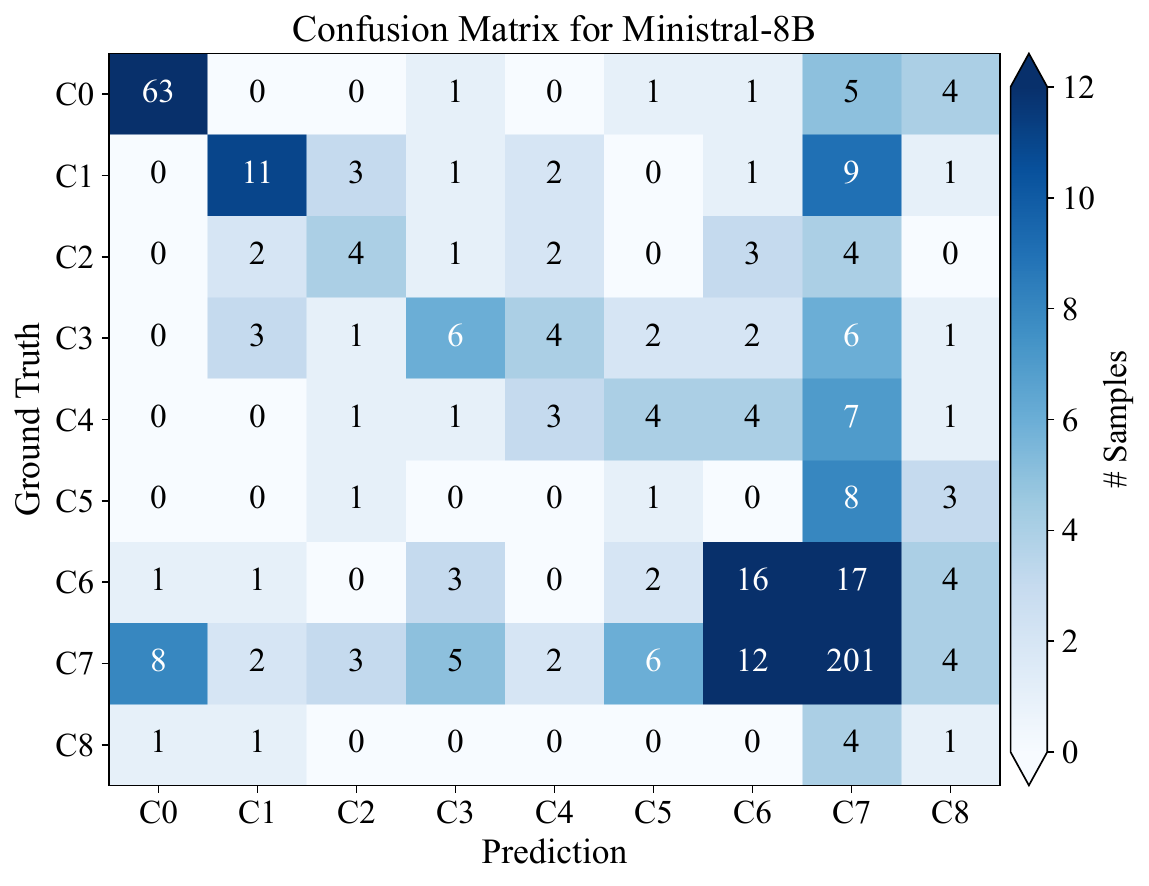}
    \caption{Confusion matrices of Gemini 2.5 Pro and Ministral-8B on the \textsc{PsyDefConv} dataset.}
    \label{fig:confusion}
    \vspace{-4mm}
\end{figure}

\subsection{Experimental Results}

Table \ref{tab:overall} summarizes the overall performance of the models on the \textsc{PsyDefConv} dataset. Among the zero-shot prompted models, Gemini 2.5 Pro and DeepSeek-V3.2 demonstrate superior performance, while Ministral-8B and InternLM3-8B emerge as the top-performing fine-tuned models. Despite these results, all models exhibit unsatisfactory effectiveness, as even the best fine-tuned macro F1-score reaches only approximately 30. This limitation is primarily due to the dataset's inherent imbalance, which poses significant challenges in accurately detecting fine-grained psychological defense mechanisms in conversations. Notably, the thinking capabilities of DeepSeek-V3.2 and Qwen3-Next both yield performance improvements, with DeepSeek-V3.2 achieving a substantial increase of approximately 40\% in accuracy and 62\% in macro F1-score, respectively.

Figures \ref{fig:per-class} and \ref{fig:confusion} illustrate the F1-scores across different psychological defense levels and the confusion matrices of the model predictions.
Both zero-shot prompted and fine-tuned models exhibit a strong bias toward predicting the \textit{high-adaptive level}, along with the action, disavowal, and minor image-distorting levels. Notably, many samples from other levels, particularly the action level and obsessional level, are frequently misclassified as high-adaptive level.
Although fine-tuning improves the model's ability to predict underrepresented levels (e.g., neurotic level), the dataset's inherent imbalance exacerbates the skew toward the dominant high-adaptive level. Furthermore, "needs more information" remains the most challenging class to predict, with Llama3.1-8B performing especially poorly, as it fails to accurately classify any sample into this category.

\section{Background and Related Work}
\subsection{Psychological Defense Mechanisms}
Psychological defense mechanisms are a cornerstone of psychodynamic theory. They refer to the unconscious processes through which the ego manages conflicts among the id, the superego, and external reality \cite{freud1936inhibitions}.
Defense mechanisms function by distorting or denying aspects of reality to reduce anxiety and protect the sense of self. While these processes are indispensable for psychological stability, their rigid or excessive use can hinder adaptation and contribute to psychological distress or interpersonal difficulties.
In supportive conversations, they often surface in language, as avoidance, rationalization, resistance, or constructive coping, shaping how the dialogue unfolds. Recognizing the defensive function of a speaker’s words can offer valuable insight into their inner state and guide more attuned emotional support.

Defense Mechanism Rating Scales (DMRS, \citealp{perry1993study}) is one of the most empirically grounded instruments for assessing defenses.
It arranges these mechanisms into a seven-level hierarchy, from Level 1 (Action defenses) to Level 7 (High Adaptive defenses), providing a structured way to evaluate overall defensive functioning based on psychotherapy transcripts or interview material \cite{perry2004studying,HierarchyDMRSQ}.
However, applying these concepts to natural language in an automated way remains an emerging challenge. The subtle, context-dependent nature of defensive expression poses significant difficulties for current NLP models. Bridging this gap will require theory-informed datasets and computational approaches sensitive to the nuances of defensive communication.

\subsection{Emotional Support Conversations}

Research on ESC aims to build dialogue agents that alleviate users’ stress through multi-turn, strategy-grounded interaction. The foundational ESConv corpus~\cite{liu-etal-2021-towards} established this direction with 1,053 dialogues and eight annotated support strategies.
The Extended ESConv and ESConv-SRA datasets~\cite{madani-srihari-2025-steering} crop long conversations and generate strategy-conditioned continuations, enabling analysis of how LLMs maintain coherence and strategy consistency with enhanced turns and strategy types.
Multi-Strategy ESConv (Dv1/Dv2)~\cite{bai2025emotionalsupportersusemultiple} emphasizes multiple strategies within a single turn and reconstructs ESConv toward 258 distinct strategy sequences, showing that LLMs outperform supervised models in producing multi-strategy replies.


Another line of work involves creating synthetic conversations to address data scarcity. \citet{zheng-etal-2023-augesc} presents AugESC, which fine-tunes GPT-J 6B on ESConv and uses it to complete dialogue threads from EmpatheticDialogues.
\citet{zheng2023buildingemotionalsupportchatbots} creates ExTES via ChatGPT’s in-context generation, which is of comparable or slightly higher quality than human-collected benchmarks. 
\citet{zhang-etal-2024-escot} introduces the ESD-CoT dataset, extending ESConv with explicit reasoning to improve interpretability.
ServeForEmo \cite{ye-etal-2025-sweetiechat} is constructed using a role-playing framework with three LLM agents: the help-seeker, the strategy-advisor, and the supporter. 


\section{Conclusion}
We study defensive functioning in supportive conversations by aligning clinical theory with what the text reveals. We annotate DMRS levels rather than mechanisms, release \textsc{PsyDefConv} with substantial agreement, and provide \textsc{DMRS Co-Pilot} to support efficient and consistent labeling. Benchmarks indicate that the task remains challenging, with clear headroom and a tendency to overpredict mature defenses. Our dataset and tools offer a reproducible basis for future work and encourage models that use richer context and theory-aware supervision. 

\section*{Limitations}

The primary limitations of our work include the language being limited to English and the skewed distribution of instances at the High-Adaptive level, reflecting the natural prevalence of psychological defense mechanisms in real-world contexts. While these limitations do not diminish our contributions, we encourage future practitioners to (1) extend the annotation to additional languages, such as Chinese, Japanese, and Spanish; and (2) generate synthetic dataset samples to mitigate the imbalance. Both directions can be effectively facilitated through the proposed \textsc{DMRS Co-Pilot} framework.


\section*{Ethical Considerations}
\paragraph{Licenses.}
We use conversations from ESConv \cite{liu-etal-2021-towards} under its stated terms. Our release contains only derived annotations, document identifiers, and code. We do not redistribute the original dialogues beyond what the ESConv license allows. Downstream users must obtain ESConv separately and agree to its conditions. Our annotations are released for research use and require compliance with the ESConv license.

\paragraph{Annotator Details.}
The two annotators are co-authors based in the United States and Australia. The within-subject timing study was conducted by a co-author based in Australia with doctoral-level training in psychology. All are adult researchers fluent in English. No additional demographic attributes such as age or gender were collected in order to reduce re-identification risk in a small sample. These activities involved only members of the research team working with de-identified, publicly available text; participation was voluntary and could be discontinued at any time. 

\paragraph{Ethics Review.}
This work involves secondary analysis of publicly available dialogue data and new annotations created by trained raters. No personal identifying information is collected or inferred. Because the content concerns mental health, we provided clear guidance on breaks and access to support resources. The protocol was reviewed by our institution and determined exempt from full ethics review since it uses public data and adds expert labels only.

\paragraph{Societal impact.}
The resource and models are intended for research on language and defensive functioning. They are not a diagnostic tool and should not be used to make clinical or legal decisions about individuals. Misuse risks include stigma, unwarranted profiling, or surveillance outside supportive contexts. The dataset is in English and reflects the domains covered by ESConv, so results may not generalize across cultures or clinical settings. We encourage users to report limitations, check for bias, and avoid normative claims about people or groups. 


\section*{Acknowledgments}
We would like to thank Shumao Yu for the insightful discussions and Zac Imel for his expert evaluation and constructive suggestions. We also gratefully acknowledge the support of the Label Studio Academic Program for providing access to the annotation platform used in this study. This work was partially supported by ARC LP 240200698.

\bibliography{custom}

\appendix
\section{Annotation Handbook: Operational Protocol}
\label{sec:appendix-handbook}

This appendix summarizes the coding protocol used to annotate defense functioning in supportive dialogues. It is designed to ensure consistency, traceability, and reproducibility. Definitions of individual defenses are omitted here.

\subsection{Scope and Units}
\begin{itemize}
\item \textbf{Target of annotation} The help seeker’s utterances only.
\item \textbf{Unit} One utterance at a time. An utterance may contain multiple sentences if they form a single turn.
\item \textbf{Context window} Use only the dialogue prior to and including the target utterance. Do not use future turns.
\item \textbf{Label set} DMRS levels 1 to 7. Two auxiliary labels are included: 0 for no defense and 8 for needs more information.
\end{itemize}

\subsection{Core Principles}
\begin{itemize}
\item \textbf{Primacy of context} Judge function with respect to the preceding dialogue.
\item \textbf{Function over form} Ask what the utterance achieves for the speaker in relation to stress or conflict.
\item \textbf{Emotion is not defense} Pure feeling statements are not defenses unless there is clear avoidance, distortion, or transformation.
\item \textbf{Acknowledge mature coping} Mark adaptive responses when they are supported by local evidence.
\end{itemize}

\subsection{Workflow}
\begin{itemize}
\item Read the prior context and the target utterance. Form a brief hypothesis of the salient stressor.
\item Decide whether the target utterance is a neutral or phatic act. If so, assign 0 and proceed to the next item.
\item If the utterance cannot be judged with the available context, assign 8.
\item Otherwise select one primary level from 1 to 7 that best explains the utterance’s function.
\item Optionally record one secondary candidate level if evidence is close.
\item Extract minimal evidence spans from the text. Write a one to two sentence rationale that links evidence to the level choice.
\item Record any uncertainties or edge cases in the notes field.
\end{itemize}

\subsection{Evidence and Rationale}
\begin{itemize}
\item Quote only what is necessary. Prefer short spans over paraphrase.
\item Ground each claim in the target utterance or its immediate context.
\item Avoid inferences about stable traits unless the dialogue explicitly supports them.
\item When evidence is weak, state why and consider label 8.
\end{itemize}

\subsection{Disambiguation Rules}
\begin{itemize}
\item \textbf{Multiple signals in one utterance} Choose the dominant function. If two are truly inseparable, prefer the lower maturity level or mark 8 when evidence is insufficient.
\item \textbf{Mature coping vs phatic closings} Simple thanks, greetings, and farewells are label 0. Rich reflections on received help may support a mature level if explicitly stated.
\item \textbf{Negative talk about others} Require signs of self esteem protection or blame shifting. Factual criticism alone is not a defense.
\item \textbf{Intellectual style} Distinguish descriptive or analytic language from efforts to keep feelings at a distance. The latter supports an obsessional level when grounded in text.
\item \textbf{Across time requirements} Signals that demand longitudinal evidence are rarely observable in short dialogues. Do not infer them without explicit cues in the current context.
\end{itemize}

\subsection{Typical Label 0 Situations}
\begin{itemize}
\item Greetings and farewells.
\item Simple thanks or brief acknowledgments such as yes, no, okay.
\item Logistical or factual questions and answers.
\item Neutral small talk and bodily states.
\item Clarifications of meaning without defensive intent.
\end{itemize}

\subsection{Quality Control}
\begin{itemize}
\item Two trained annotators label independently and blind to each other.
\item Use a shared handbook and examples for calibration. Hold regular reviews to refine decision rules.
\item Compute agreement statistics on double labeled items. Resolve disagreements by consensus to create gold labels.
\item Maintain an audit trail with conversation id, utterance id, primary level, secondary level when used, stressor hypothesis, evidence spans, rationale, and notes.
\end{itemize}

\subsection{Use with \textsc{DMRS Co-Pilot}}
\begin{itemize}
\item Pre annotations include a stressor hypothesis, candidate items, evidence summaries, and two ranked level suggestions.
\item Annotators must verify or revise suggestions based on the text. Record whether the suggestion was helpful.
\item When suggestions overreach beyond local evidence, prefer a conservative label or 8.
\end{itemize}

\label{sec:appendix} 

\section{Further Data Analysis}
\begin{figure}[t]
    \centering
    \includegraphics[width=1\linewidth]{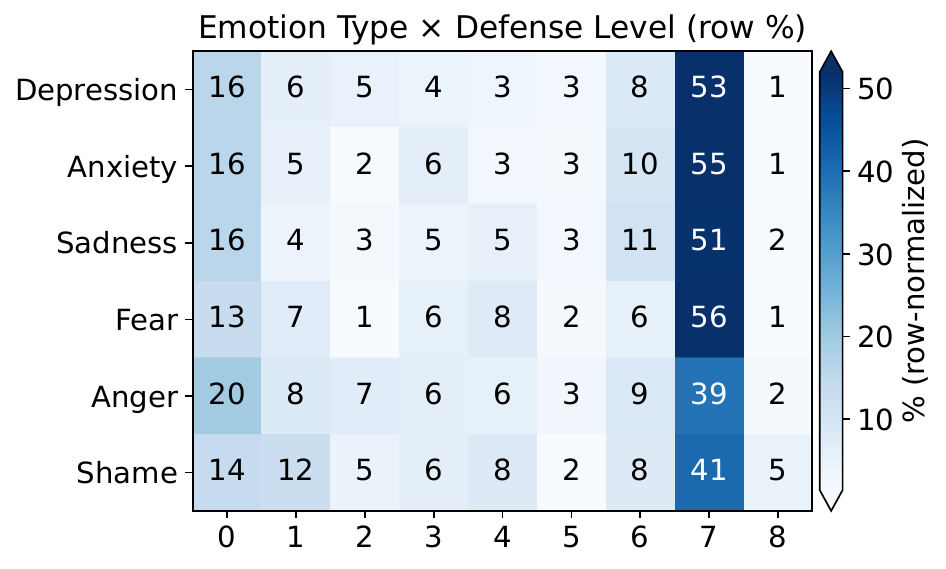}
    \caption{Defense levels by emotion. Level 7 dominates. Shame shows more low levels. Anger shows more Level 0. Depression anxiety and sadness are similar.}
    \label{fig:app_heatmap_emotion}
    \vspace{-1mm}
\end{figure}

\begin{figure}[t]
    \centering
    \includegraphics[width=1\linewidth]{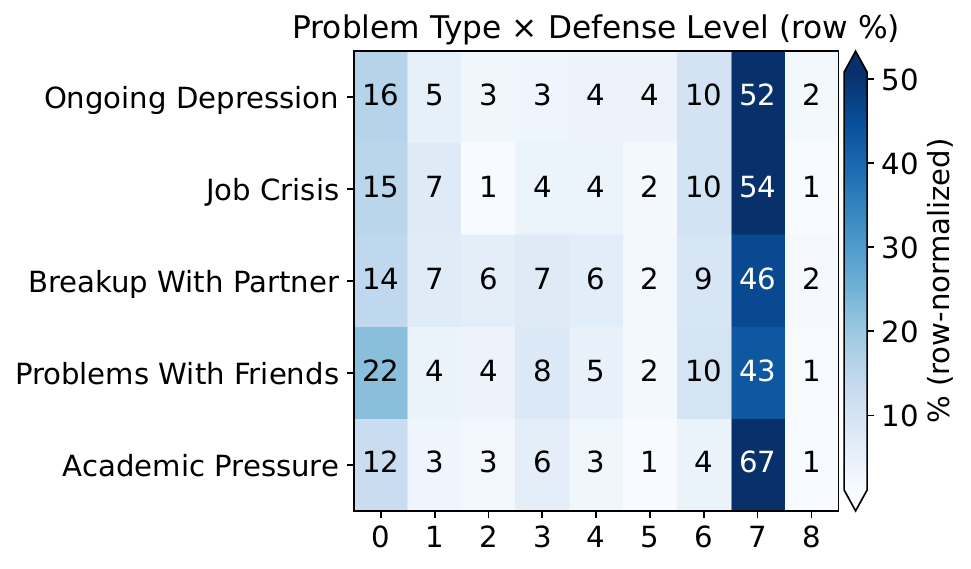}
    \caption{Defense levels by problem context. Level 7 dominates. Academic pressure is highest at Level 7. Problems with friends shows more Level 0. Breakup with partner is higher at Levels 2 to 4.}
    \label{fig:app_heatmap_problem}
    \vspace{-4mm}
\end{figure}

\paragraph{Heatmaps of Defense Distributions.}
Figure~\ref{fig:app_heatmap_emotion} shows that mature defenses (Level~7) dominate across all emotions, indicating a general preference for adaptive coping. However, variations emerge across emotion types. Shame exhibits the highest proportion of low-level defenses (Levels~1–4), suggesting more fragmented and immature regulation. Anger stands out for its elevated use of Level~0 (no defense), implying more unfiltered or confrontational expressions. In contrast, \textit{depression}, \textit{anxiety}, and \textit{sadness} exhibit highly similar distributions across all nine defense levels, reflecting a consistent and balanced defensive response. Turning to problem contexts, Figure~\ref{fig:app_heatmap_problem} reports row normalized percentages that mirror the dominance of Level~7 across contexts, with academic pressure showing the largest share at Level~7 and the smallest shares at low levels, problems with friends showing the highest share at Level~0, breakup with partner concentrating more mass in Levels~2 to 4 than other contexts, while ongoing depression and job crisis align closely with the overall pattern.

\begin{figure*}[t]
    \centering
    \includegraphics[width=1\linewidth]{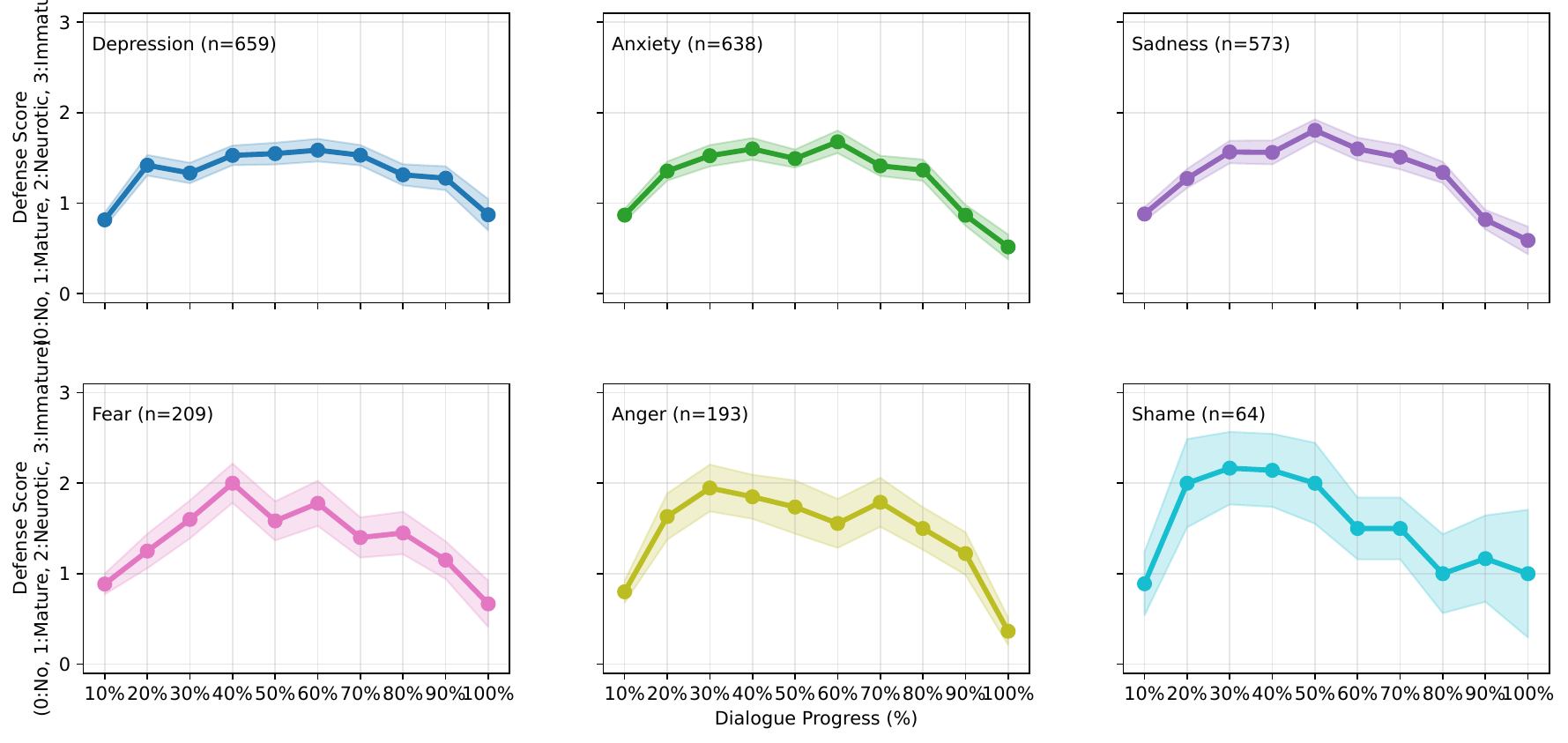}
    \vspace{-5pt}
    \caption{Defense score trajectories across dialogue progress by emotion. Each curve shows the mean score with a shaded band that indicates variability. Higher scores indicate more immature and neurotic defenses and lower scores indicate mature or no defense. Shame rises early then declines. Anger ends near the lowest scores. Sadness depression and anxiety follow closely aligned paths.}
    \label{fig:app_progress_emotions}
    \vspace{-15pt}
\end{figure*}

\begin{figure*}[t]
    \centering
    \includegraphics[width=1\linewidth]{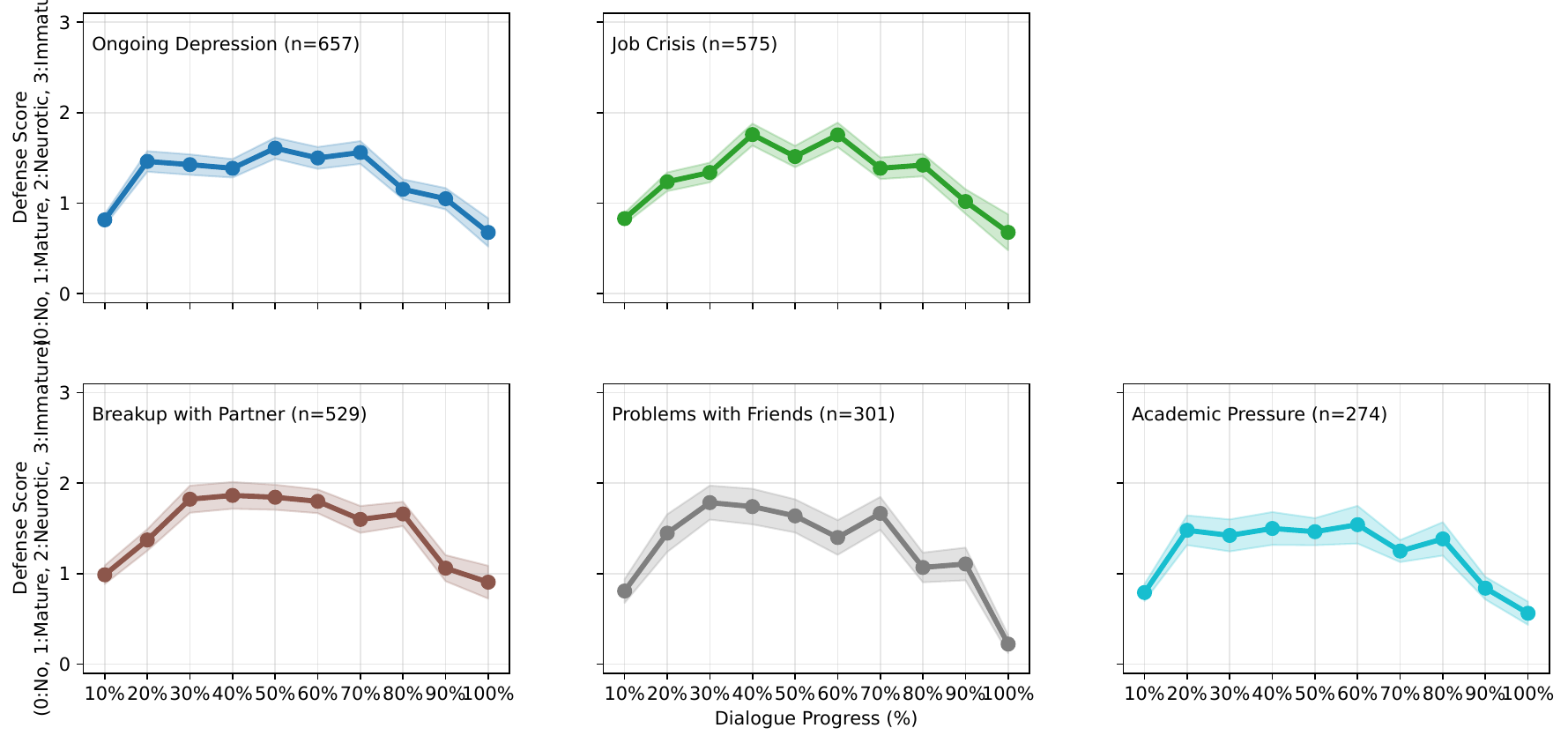}
    \vspace{-5pt}
    \caption{Defense score trajectories across dialogue progress by problem context. Most contexts peak in the middle then return to lower scores at the end. Job crisis shows the clearest arch. Breakup with partner rises early then holds a plateau before easing. Ongoing depression and problems with friends show moderate peaks. Academic pressure varies little and ends lower than the middle.}
    \label{fig:app_progress_problems}
    \vspace{-15pt}
\end{figure*}

\paragraph{Progress of Defense Strategies Across Emotions and Problem Contexts.}
Figure~\ref{fig:app_progress_emotions} and Figure~\ref{fig:app_progress_problems} trace average defense scores across dialogue progress from 10 percent to 100 percent. On the vertical axis larger values indicate more immature and neurotic defenses while lower values indicate mature or no defense. Across emotions and problem contexts the curves generally rise in the early to middle stages then return toward more regulated responding by the end. Shame shows the sharpest early ascent followed by a clear decline. Sadness depression and anxiety follow similar and smooth paths near mature defenses. Anger moves toward the lowest scores near the closing turns which suggests more direct expression with minimal defensive filtering. By context job crisis shows the most pronounced arch with a high middle peak. Ongoing depression and problems with friends peak in the middle with moderate amplitude. Breakup with partner climbs early then holds a plateau before easing. Academic pressure varies the least and ends with a clear return toward mature defenses. These patterns highlight the middle of the dialogue as a critical window for defense regulation and complement the findings in the main text.

\section{HumanSignal Annotation Interface}
\begin{figure*}[htbp]
    \centering
    \includegraphics[width=1\linewidth]{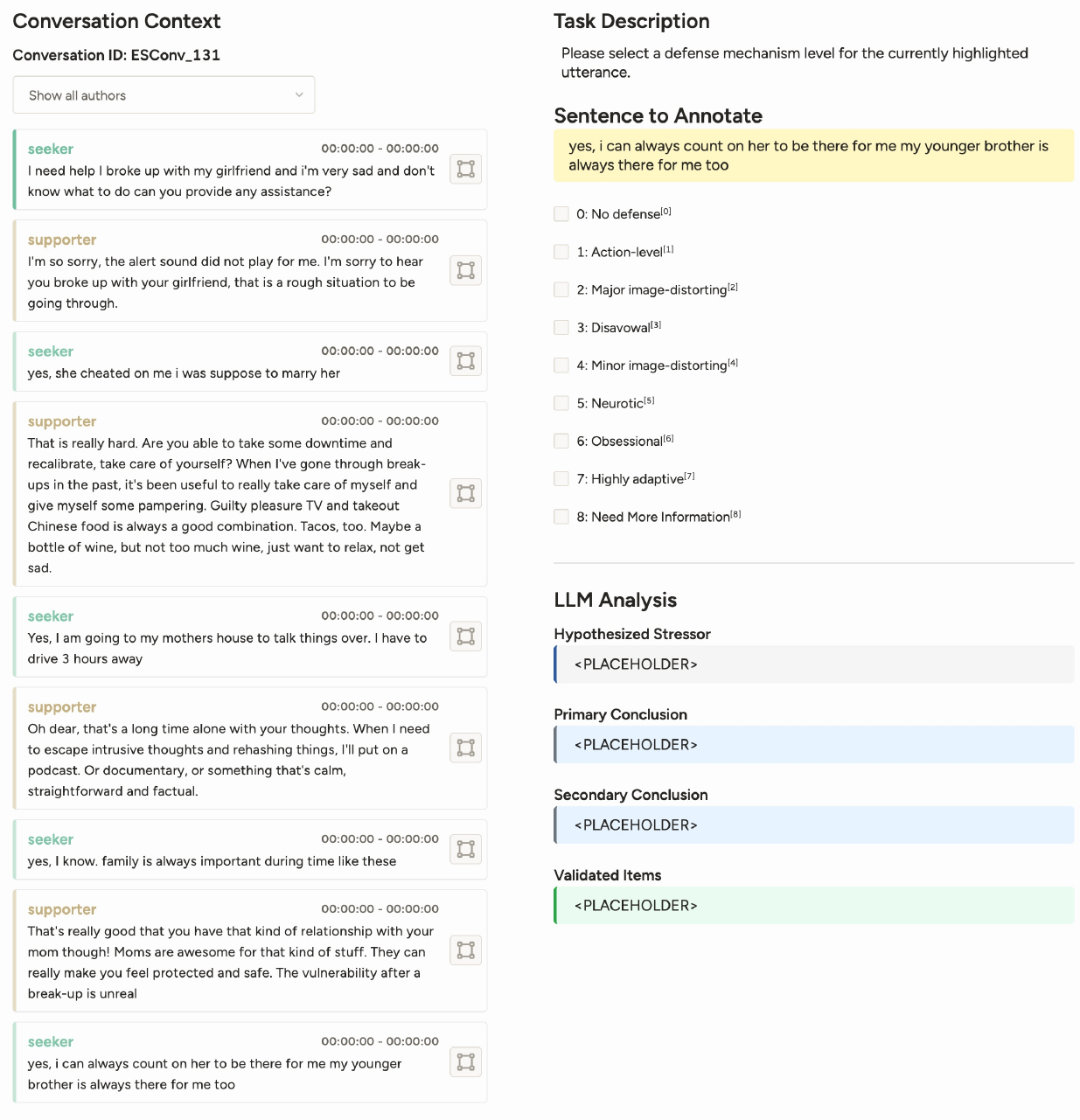}
    \vspace{-5pt}
    \caption{Humansignal interface for defense level annotation. Left column shows the full dialogue. Right panel prompts the rater to select a level from 0 to 8 and provides fields for analysis and validation.}
    \label{fig:app_humansignal}
    \vspace{-15pt}
\end{figure*}

Figure~\ref{fig:app_humansignal} presents the Humansignal annotation screen on page 1. The left column shows the conversation context with seeker and supporter turns. The right panel states the task and highlights the sentence to annotate. Annotators choose a defense level from 0 to 8 where 0 is no defense and 7 is highly adaptive while 8 denotes need more information. The panel also includes structured fields for hypothesized stressor primary conclusion secondary conclusion and validated items. This figure documents the data collection workflow used to generate defense labels and complements the quantitative figures by showing the exact labeling environment.

\section{DMRS Co-Pilot System Prompts}
The detailed prompts will be made publicly available upon the publication of this paper.

\end{document}